\documentclass[11pt,a4paper]{article}
\usepackage[hyperref]{naaclhlt2018}
\usepackage{times}
\usepackage{latexsym}
\usepackage{subcaption}
\usepackage{graphicx}
\usepackage{tabularx}
\usepackage{multicol}
\usepackage{colortbl}
\usepackage{makecell}
\usepackage{amssymb}
\usepackage{multirow}
\usepackage{amsmath}
\usepackage{tikz}

\DeclareCaptionFont{10pt}{\fontsize{10pt}{12pt}\selectfont}
\usepackage{url}

\aclfinalcopy 

\newcolumntype{Y}{>{\centering\let\newline\\\arraybackslash\hspace{0pt}}X}

\title{Entity Commonsense Representation\\ for Neural Abstractive Summarization}

\author{Reinald Kim Amplayo\thanks{{ } Amplayo and Lim are co-first authors with equal contribution. Names are arranged alphabetically.} \and
  Seonjae Lim\footnotemark[1] \and
  Seung-won Hwang \\
  Yonsei University, Seoul, South Korea \\
  {\tt \{rktamplayo, sun.lim, seungwonh\}@yonsei.ac.kr} \\
}

\date{}

\begin{document}
\maketitle
\begin{abstract}
  A major proportion of a text summary includes important entities found in the original text. These entities build up the topic of the summary. Moreover, they hold commonsense information once they are linked to a knowledge base. Based on these observations, this paper investigates the usage of linked entities to guide the decoder of a neural text summarizer to generate concise and better summaries. To this end, we leverage on an off-the-shelf entity linking system (ELS) to extract linked entities and propose \textbf{Entity2Topic (E2T)}, a module easily attachable to a sequence-to-sequence model that transforms a list of entities into a vector representation of the topic of the summary. Current available ELS's are still not sufficiently effective, possibly introducing unresolved ambiguities and irrelevant entities. We resolve the imperfections of the ELS by (a) encoding entities with selective disambiguation, and (b) pooling entity vectors using firm attention. By applying E2T to a simple sequence-to-sequence model with attention mechanism as base model, we see significant improvements of the performance in the Gigaword (sentence to title) and CNN (long document to multi-sentence highlights) summarization datasets by at least 2 ROUGE points.
\end{abstract}

\section{Introduction}
\label{sec:intro}

Text summarization is a task to generate a shorter and concise version of a text while preserving the meaning of the original text. The task can be divided into two subtask based on the approach: extractive and abstractive summarization. Extractive summarization is a task to create summaries by pulling out snippets of text form the original text and combining them to form a summary. Abstractive summarization
asks to generate summaries from scratch without the restriction to use the available words from the original text. Due to the limitations of extractive summarization on incoherent texts and unnatural methodology \cite{yao2017recent}, the research trend has shifted towards abstractive summarization.

Sequence-to-sequence models \cite{sutskever2014sequence} with attention mechanism \cite{bahdanau2014neural} have found great success in generating abstractive summaries, both from a single sentence \cite{chopra2016abstractive} and from a long document with multiple sentences \cite{chen2016distraction}. However, when generating summaries, it is necessary to determine the main topic and to sift out unnecessary information that can be omitted. Sequence-to-sequence models have the tendency to include all the information, relevant or not, that are found in the original text. This may result to unconcise summaries that concentrates wrongly on irrelevant topics. The problem is especially severe when summarizing longer texts.


\begin{figure}
    \centering
    \includegraphics[width=0.47\textwidth]{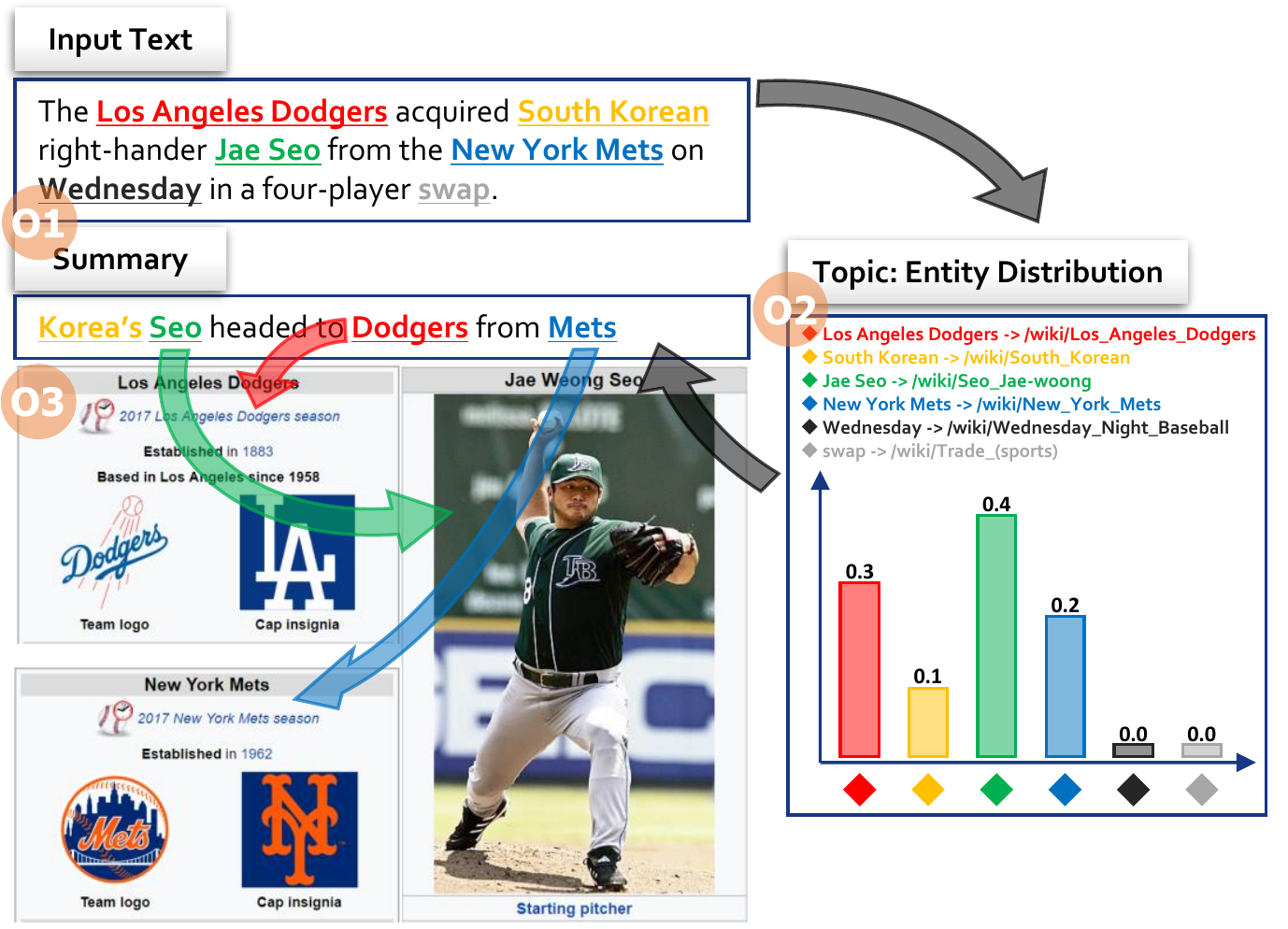}
    \caption{Observations on linked entities in summaries. \textbf{O1}: Summaries are mainly composed of entities. \textbf{O2}: Entities can be used to represent the topic of the summary. \textbf{O3}: Entity commonsense learned from a large corpus can be used.}
    \label{fig:intuition}
\end{figure}

In this paper, we propose to use entities found in the original text to infer the summary topic, mitigating the aforementioned problem.
Specifically, we leverage on linked entities extracted by employing
a readily available entity linking system.
The importance of using linked entities in summarization is intuitive and can be explained by looking at Figure~\ref{fig:intuition} as an example. First (\textbf{O1} in the Figure), aside from auxiliary words to construct a sentence, a summary is mainly composed of linked entities extracted from the original text. Second (\textbf{O2}), we can depict the main topic of the summary as a probability distribution of relevant entities from the list of entities. Finally (\textbf{O3}), we can leverage on entity commonsense learned from a separate large knowledge base such as Wikipedia.

To this end, we present a method to effectively apply linked entities in sequence-to-sequence models, called \textbf{Entity2Topic (E2T)}. E2T is a module that can be easily attached to any sequence-to-sequence based summarization model. The module encodes the entities extracted from the original text by an entity linking system (ELS), constructs a vector representing the topic of the summary to be generated, and informs the decoder about the constructed topic vector. Due to the imperfections of current ELS's, the extracted linked entities may be too \textbf{ambiguous} and \textbf{coarse} to be considered relevant to the summary. We solve this issue by using entity encoders with \textbf{selective disambiguation} and by constructing topic vectors using \textbf{firm attention}.

We experiment on two datasets, Gigaword and CNN, with varying lengths. We show that applying our module to a sequence-to-sequence model with attention mechanism significantly increases its performance on both datasets.
Moreover, when compared with the state-of-the-art models for each dataset, the model obtains a comparable performance on the Gigaword dataset where the texts are short, and outperforms all competing models on the CNN dataset where the texts are longer.
Furthermore, we provide analysis on how our model effectively uses the extracted linked entities to produce concise and better summaries.

\section{Usefulness of linked entities in summarization}
\label{sec:ent4sum}

In the next subsections, we present detailed arguments with empirical and previously examined evidences on the observations and possible issues when using linked entities extracted by an entity linking system (ELS) for generating abstractive summaries. For this purpose, we use the development sets of the Gigaword dataset provided in \cite{rush2015neural} and of the CNN dataset provided in \cite{hermann2015teaching} as the experimental data for quantitative evidence and refer the readers to Figure \ref{fig:intuition} as the running example.

\subsection{Observations}

As discussed in Section \ref{sec:intro}, we find three observations that show the usefulness of linked entities for abstractive summarization.

First, summaries are mainly composed of linked entities extracted from the original text. In the example, it can be seen that the summary contains four words that refer to different entities.
In fact, all noun phrases in the summary mention at least one linked entity. In our experimental data, we extract linked entities from the original text and compare them to the noun phrases found in the summary. We report that $77.1\%$ and $75.1\%$ of the noun phrases on the Gigaword and CNN datasets, respectively, contain at least one linked entity, which confirms our observation.

Second, linked entities can be used to represent the topic of the summary,
defined as a multinomial distribution over entities, as graphically shown in the example, where the probabilities refer to the relevance of the entities. Entities have been previously used to represent topics
\cite{newman2006analyzing}, as they can be utilized as a controlled vocabulary of the main topics in a document \cite{hulpus2013unsupervised}. 
In the example, we see that the entity ``\textit{Jae Seo}'' is the most relevant because it is the subject of the summary, while the entity ``\textit{South Korean}'' is less relevant because it is less important when constructing the summary.

Third, we can make use of the entity commonsense that can be learned as a continuous vector representation from a separate larger corpus \cite{ni2016semantic,yamada2017learning}. In the example, if we know that the entities ``\textit{Los Angeles Dodgers}'' and ``\textit{New York Mets}'' are American baseball teams and ``\textit{Jae Seo}'' is a baseball player associated with the teams, then we can use this information to generate more coherent summaries. 
We find that $76.0\%$ of the extracted linked entities are covered by the pre-trained vectors\footnote{\url{https://github.com/idio/wiki2vec}} in our experimental data, proving our third observation.

\subsection{Possible issues}
\label{sec:issues}

Despite its usefulness, linked entities extracted from ELS's 
have issues because of low precision rates
\cite{hasibi2016reproducibility} and design challenges in training datasets \cite{ling2015design}. These issues can be summarized into two parts: ambiguity and coarseness.

First, the extracted entities may be ambiguous. In the example, the entity ``\textit{South Korean}'' is ambiguous because it can refer to both the South Korean person and the South Korean language, among others\footnote{\url{https://en.wikipedia.org/wiki/South_Korean}}.
In our experimental data, we extract (1) the top 100 entities based on frequency, and (2) the entities extracted from 100 randomly selected texts, and check whether they have disambiguation pages in Wikipedia or not. We discover that $71.0\%$ of the top 100 entities and $53.6\%$ of the entities picked at random have disambiguation pages, which shows that most entities are prone to ambiguity problems.

Second, the linked entities may also be too common to be considered an entity.
This may introduce \textit{errors} and \textit{irrelevance} to the summary.
In the example, ``\textit{Wednesday}'' is erroneous because it is wrongly linked to the entity ``\textit{Wednesday Night Baseball}''. Also, ``\textit{swap}'' is irrelevant because although it is linked correctly to the entity ``\textit{Trade (Sports)}'', it is too common and irrelevant when generating the summaries. In our experimental data, we randomly select 100 data instances and tag the correctness and relevance of extracted entities into one of four labels: A: correct and relevant, B: correct and somewhat relevant, C: correct but irrelevant, and D: incorrect. Results show that $29.4\%$, $13.7\%$, $30.0\%$, and $26.9\%$ are tagged with A, B, C, and D, respectively, which shows that there is a large amount of incorrect and irrelevant entities.

\begin{figure*}[t]
    \centering
    \includegraphics[width=0.95\textwidth]{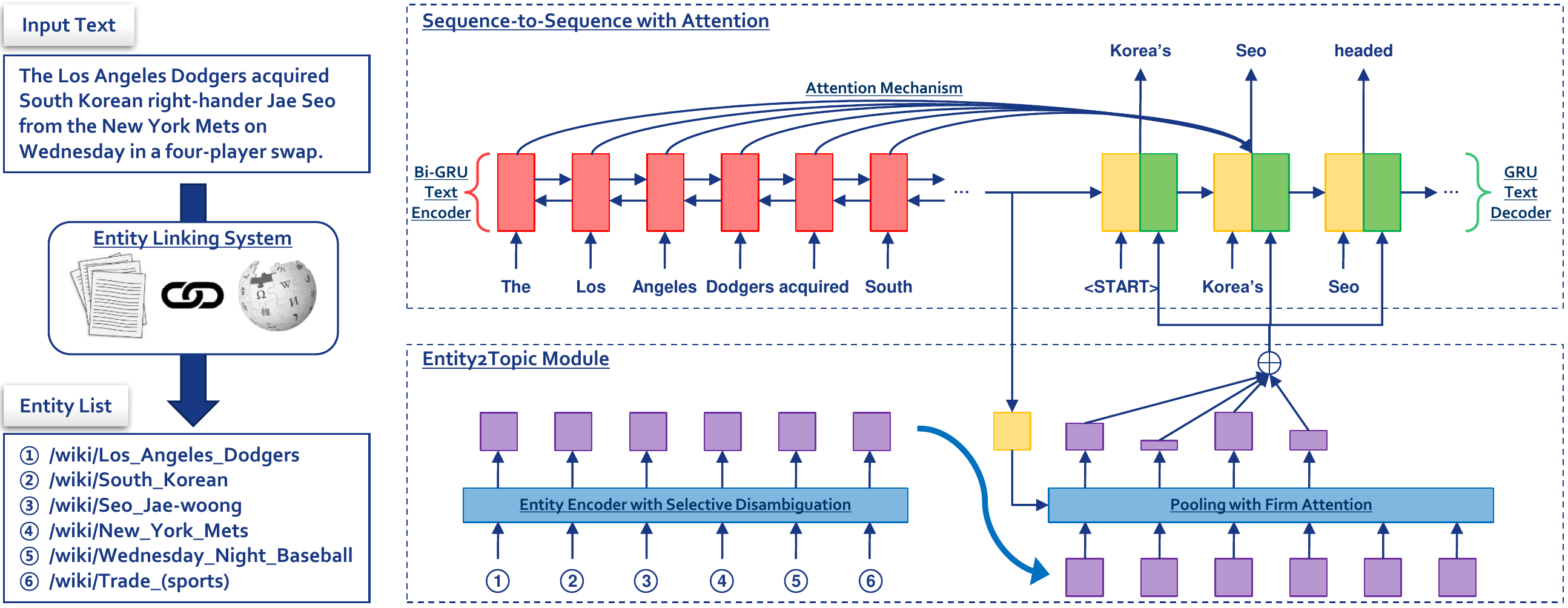}
    \caption{Full architecture of our proposed sequence-to-sequence model with Entity2Topic (E2T) module.}
    \label{fig:model}
\end{figure*}

\section{Our model}

To solve the issues described above, we present \textbf{Entity2Topic (E2T)}, a module that can be easily attached to any sequence-to-sequence based abstractive summarization model. E2T encodes the linked entities extracted from the text and transforms them into a single topic vector. This vector is ultimately concatenated to the decoder hidden state vectors. The module contains two submodules specifically for the issues presented by the entity linking systems: the entity encoding submodule with selective disambiguation and the pooling submodule with firm attention.

Overall, our full architecture can be illustrated as in Figure \ref{fig:model}, which consists of an entity linking system (ELS), a sequence-to-sequence with attention mechanism model, and the E2T module.
We note that our proposed module can be easily attached to more sophisticated abstractive summarization models \cite{zhou2017selective,tan2017abstractive} that are based on the traditional encoder-decoder framework and consequently can produce better results.
The code of the base model and the E2T are available online\footnote{\url{https://github.com/rktamplayo/Entity2Topic}}.

\subsection{Base model}

As our base model, we employ a basic encoder-decoder RNN used in most neural machine translation \cite{bahdanau2014neural} and text summarization \cite{nallapati2016abstractive} tasks. We employ a two-layer bidirectional GRU (BiGRU) as the recurrent unit of the encoder.
%
%
The BiGRU consists of a forward and backward GRU, which results to sequences of forward and backward hidden states $(\overrightarrow{h}_1, \overrightarrow{h}_2, ..., \overrightarrow{h}_n)$ and $(\overleftarrow{h}_1, \overleftarrow{h}_2, ..., \overleftarrow{h}_n)$, respectively:
\begin{align*}
    \overrightarrow{h}_i &= GRU(x_i, \overrightarrow{h}_{i-1}) \\
    \overleftarrow{h}_i &= GRU(x_i, \overleftarrow{h}_{i+1}) \nonumber
\end{align*}

The forward and backward hidden states are concatenated to get the hidden state vectors of the tokens (i.e. $h_i = [\overrightarrow{h}_i; \overleftarrow{h}_i]$). The final states of the forward and backward GRU are also concatenated to create the final text representation vector of the encoder $s = [\overrightarrow{h}_n; \overleftarrow{h}_1]$. These values are calculated per layer, where $x_t$ of the second layer is $h_t$ of the first layer. The final text representation vectors are projected by a fully connected layer and are passed to the decoder as the initial hidden states $s_0 = s$.

For the decoder, we use a two-layer uni-directional GRU with attention. At each time step $t$, the previous token $y_{t-1}$, the previous hidden state $s_{t-1}$, and the previous context vector $c_{t-1}$ are passed to a GRU to calculate the new hidden state $s_t$, as shown in the equation below.
\begin{equation}
    s_t = GRU(w_{t-1}, s_{t-1}, c_{t-1}) \nonumber
\end{equation}

The context vector $c_t$ is computed using the additive attention mechanism \cite{bahdanau2014neural}, which matches the current decoder state $s_t$ and each encoder state $h_i$ to get an importance score. The scores are then passed to a softmax and are used to pool the encoder states using weighted sum. The final pooled vector is the context vector, as shown in the equations below.
\begin{align*}
    g_{t,i} &= v_a^\top tanh(W_a s_{t-1} + U_a h_i) \\
    a_{t,i} &= \frac{exp(g_{t,i})}{\sum_i exp(g_{t,i})} \\
    c_t &= \sum_i a_{t,i} h_i \nonumber
\end{align*}

Finally, the previous token $y_{t-1}$, the current context vector $c_t$, and the current decoder state $s_t$ are used to generate the current word $y_t$ with a softmax layer over the decoder vocabulary, as shown below.
\begin{align*}
    o_t &= W_w w_{t-1} + W_c c_t + W_s s_t \\ 
    p(y_t | y_{<t}) &= softmax(W_o o_t) \nonumber
\end{align*}

\subsection{Entity encoding submodule}

After performing entity linking to the input text using the ELS, we receive a sequential list of linked entities, arranged based on their location in the text. We embed these entities to $d$-dimensional vectors $E = \{e_1, e_2, ..., e_m\}$ where $e_i \in \mathbb{R}^d$. Since these entities may still contain ambiguity, it is necessary to resolve them before applying them to the base model. Based on the idea that an ambiguous entity can be disambiguated using its neighboring entities, we introduce two kinds of disambiguating encoders below.

\paragraph{Globally disambiguating encoder}

One way to disambiguate an entity is by using all the other entities, putting more importance to entities that are nearer. For this purpose, we employ an RNN-based model to globally disambiguate the entities. Specifically, we use BiGRU and concatenate the forward and backward hidden state vectors as the new entity vector:
\begin{align*}
    \overrightarrow{h}_i &= GRU(e_i, \overrightarrow{h}_{i-1}) \\
    \overleftarrow{h}_i &= GRU(e_i, \overleftarrow{h}_{i+1}) \\
    e'_i &= [\overrightarrow{h}_i; \overleftarrow{h}_i] \nonumber
\end{align*}

\paragraph{Locally disambiguating encoder}

Another way to disambiguate an entity is by using only the direct neighbors of the entity, putting no importance value to entities that are far. To do this, we employ a CNN-based model to locally disambiguate the entities. Specifically, we do the convolution operation using filter matrices $W_f \in \mathbb{R}^{h \times d}$ with filter size $h$ to a window of $h$ words. We do this for different sizes of $h$. This produces new feature vectors $c_{i,h}$ as shown below, where $f(.)$ is a non-linear function:
\begin{equation*}
    c_{i,h} = f([e_{i-(h-1)/2}; ...; e_{i+h(+1)/2}]^\top W_f + b_f) \nonumber
\end{equation*}

The convolution operation reduces the number of entities differently depending on the filter size $h$. To prevent loss of information and to produce the same amount of feature vectors $c_{i,h}$, we pad the entity list dynamically such that when the filter size is $h$, the number of paddings on each side is $(h-1)/2$. The filter size $h$ therefore refers to the number of entities used to disambiguate a middle entity. Finally, we concatenate all feature vectors of different $h$'s for each $i$ as the new entity vector:
\begin{equation*}
    e'_i = [c_{i,h_1}; c_{i, h_2}; ...] \nonumber
\end{equation*}

The question on which disambiguating encoder is better has been a debate; some argued that using only the local context is appropriate \cite{lau2013unimelb} while some claimed that additionally using global context also helps \cite{wang2015sense}.
The RNN-based encoder is good as it smartly makes use of all entities, however it may perform bad when there are many entities as it introduces noise when using a far entity during disambiguation. The CNN-based encoder is good as it minimizes the noise by totally ignoring far entities when disambiguating, however determining the appropriate filter sizes $h$ needs engineering. Overall, we argue that when the input text is short (e.g. a sentence), both encoders perform comparably, otherwise when the input text is long (e.g. a document), the CNN-based encoder performs better.

\paragraph{Selective disambiguation}

\begin{figure}[t]
    \centering
    \includegraphics[width=0.47\textwidth]{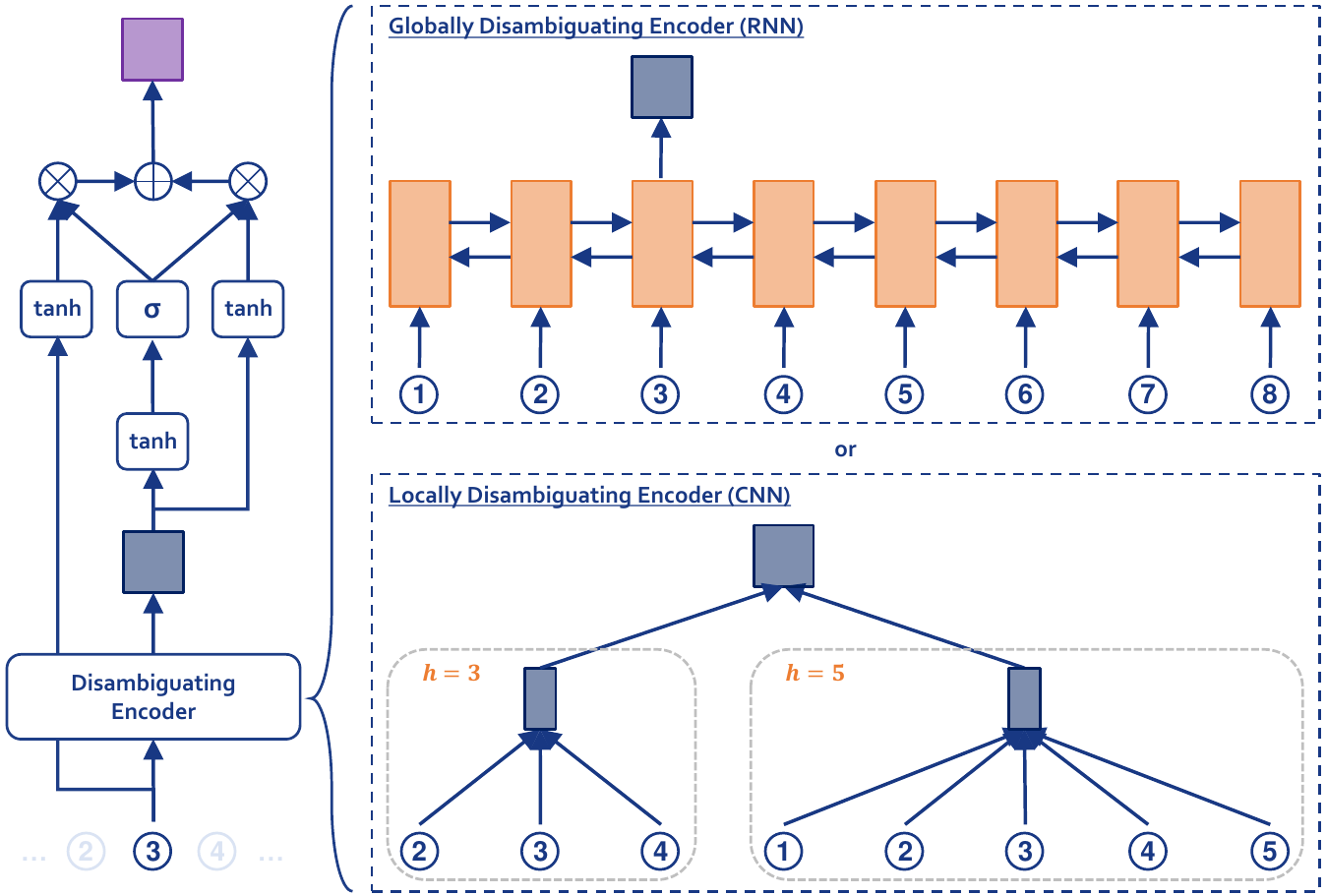}
    \caption{Entity encoding submodule with selective disambiguation applied to the entity \textcircled{\raisebox{-0.9pt}{3}}. 
    The left figure represents the full submodule while the right figure represents the two choices of disambiguating encoders.
    }
    \label{fig:sd}
\end{figure}

It is obvious that not all entities need to be disambiguated. When a correctly linked and already adequately disambiguated entity is disambiguated again, it would make the entity very context-specific and might not be suitable for the summarization task. Our entity encoding submodule therefore uses a selective mechanism that decides whether to use the disambiguating encoder or not. This is done by introducing a selective disambiguation gate $d$. The final entity vector $\tilde{e}_i$ is calculated as the linear transformation of $e_i$ and $e'_i$:
\begin{align*}
    e'_i &= encoder(e_i) \\
    d &= \sigma(W_d e'_i + b_d) \\
    \tilde{e}_i &= d \times f(W_x e_i + b_x) + \\ & \quad (1-d) \times f(W_y e'_i + b_y) \nonumber
\end{align*}

The full entity encoding submodule is illustrated in Figure \ref{fig:sd}. Ultimately, the submodule outputs the disambiguated entity vectors $\tilde{E} = \{\tilde{e}_1, \tilde{e}_2, ..., \tilde{e}_m\}$.

\subsection{Pooling submodule}

The entity vectors $\tilde{E}$ are pooled to create a single topic vector $t$ that represents the topic of the summary. One possible 
pooling technique 
is to use soft attention \cite{xu2015show} on the vectors to determine the importance value of each vector, which can be done by matching each entity vector with the text vector $s$ from the text encoder as the context vector. The entity vectors are then pooled using weighted sum.
%
One problem with soft attention is that it considers \textit{all} entity vectors when constructing the topic vector. However, not all entities are important and necessary when generating summaries. Moreover, a number of these entities may be erroneous and irrelevant, as reported in Section \ref{sec:issues}. Soft attention gives non-negligible important scores to these entities, thus adds unnecessary noise to the construction of the topic vector.

Our pooling submodule instead uses \textbf{firm attention} mechanism to consider only top $k$ entities when constructing the topic vector. This is done in a differentiable way as follows:
\begin{align*}
    G &= v_a^\top tanh(W_a \tilde{E} + U_a s) \\
    K &= top\_k(G) \\
    P &= sparse\_vector(K, 0, -\infty) \\
    g'_i &= g_i + p_i \\
    a_i &= \frac{exp(g'_i)}{\sum_i exp(g'_i)} \\
    t &= \sum_i a_i \tilde{e}_i \nonumber
\end{align*}
where the functions $K = top\_k(G)$ gets the indices of the top $k$ vectors in $G$ and $P = sparse\_vector(K,0,-\infty)$ creates a sparse vector where the values of $K$ is $0$ and $-\infty$ otherwise\footnote{We use $-10^9$ to represent $-\infty$.}. The sparse vector $P$ is added to the original importance score vector $G$ to create a new importance score vector. In this new vector, important scores of non-top $k$ entities are $-\infty$.
When softmax is applied, this gives very small, negligible, and close-to-zero values to non-top $k$ entities.
The value $k$ depends on the lengths of the input text and summary.
Moreover, when $k$ increases towards infinity, firm attention becomes soft attention.
We decide $k$ empirically (see Section \ref{sec:exp}).

\subsection{Extending from the base model}

Entity2Topic module extends the base model as follows. The final text representation vector $s$ is used as a context vector when constructing the topic vector $t$ in the pooling submodule. The topic vector $t$ is then concatenated to the decoder hidden state vectors $s_i$, i.e. $s'_i = [s_i; t]$. The concatenated vector is finally used to create the output vector:
\begin{equation}
    o_i = W_w w_{i-1} + W_c c_i + W_s s'_i \nonumber
\end{equation}

\section{Related work}


Due to its recent success, neural network models have been used with competitive results on abstractive summarization. A neural attention model was first applied to the task, easily achieving state-of-the-art performance on multiple datasets \cite{rush2015neural}. The model has been extended to instead use recurrent neural network as decoder \cite{chopra2016abstractive}. The model was further extended to use a full RNN encoder-decoder framework and further enhancements through lexical and statistical features \cite{nallapati2016abstractive}. The current state-of-the-art performance is achieved by selectively encoding words as a process of distilling salient information \cite{zhou2017selective}.

Neural abstractive summarization models have also been explored to summarize longer documents. Word extraction models
have been previously explored, performing worse than sentence extraction models \cite{cheng2016neural}. Hierarchical attention-based recurrent neural networks have also been applied to the task, owing to the idea that there are multiple sentences in a document \cite{nallapati2016abstractive}. Finally, distraction-based models were proposed to enable models to traverse the text content and grasp the overall meaning \cite{chen2016distraction}. The current state-of-the-art performance is achieved by a graph-based attentional neural model, considering the key factors of document summarization such as saliency, fluency and novelty \cite{tan2017abstractive}.




Previous studies on the summarization tasks have only used entities in the preprocessing stage to anonymize the dataset \cite{nallapati2016abstractive} and to mitigate out-of-vocabulary problems \cite{tan2017abstractive}.
Linked entities for summarization are still not properly explored and we are the first to use linked entities to improve the performance of the summarizer.

\section{Experimental settings}
\label{sec:exp}

\paragraph{Datasets}

We use two widely used summarization datasets with different text lengths. First, we use the Annotated English Gigaword dataset as used in \cite{rush2015neural}.
This dataset receives the first sentence of a news article as input and use the headline title as the gold standard summary.
Since the development dataset is large, we randomly selected 2000 pairs as our development dataset. We use the same held-out test dataset used in \cite{rush2015neural} for comparison.
Second, we use the CNN dataset released in \cite{hermann2015teaching}.
This dataset receives the full news article as input and use the human-generated multiple sentence highlight as the gold standard summary.
The original dataset has been modified and preprocessed specifically for the document summarization task \cite{nallapati2016abstractive}.
In addition to the previously provided datasets, we extract linked entities using Dexter\footnote{\url{http://dexter.isti.cnr.it/}} \cite{ceccarelli2013dexter}, an open source ELS that links text snippets found in a given text to entities contained in Wikipedia. 
We use the default recommended parameters stated in the website.
We summarize the statistics of both datasets in Table \ref{tab:statistics}.

\begin{table}[t]
  \centering
    \begin{tabular}{|c|cc|}
    \hline
    Dataset & Gigaword & CNN \\
    \hline
    num(data) & 4.0M  & 84K \\
    avg(inputWord) & 31.4  & 774.9 \\
    avg(outputWord) & 8.2   & 48.1 \\
    \hline
    min(inputEntity) & 1     & 1 \\
    max(inputEntity) & 36    & 743 \\
    avg(inputEntity) & 4.5   & 94.6 \\
    \hline
    \end{tabular}%
  \caption{Dataset statistics.}
  \label{tab:statistics}%
\end{table}%

\paragraph{Implementation}

For both datasets, we further reduce the size of the input, output, and entity vocabularies to at most 50K as suggested in \cite{see2017get} and replace less frequent words to ``$<$unk$>$''. We use 300D Glove\footnote{\url{https://nlp.stanford.edu/projects/glove/}} \cite{pennington2014glove} and 1000D wiki2vec\footnote{\url{https://github.com/idio/wiki2vec}} pre-trained vectors to initialize our word and entity vectors. For GRUs, we set the state size to 500. For CNN, we set $h=3, 4, 5$ with $400, 300, 300$ feature maps, respectively. For firm attention, $k$ is tuned
by calculating the perplexity of the model starting with smaller values (i.e. $k=1,2,5,10,20,...$) and stopping when the perplexity of the model becomes worse than the previous model.
Our preliminary tuning showed that $k=5$ for Gigaword dataset and $k=10$ for CNN dataset are the best choices.
We use dropout \cite{srivastava2014dropout} on all non-linear connections with a dropout rate of 0.5. 
We set the batch sizes of Gigaword and CNN datasets to 80 and 10, respectively.
Training is done via stochastic gradient descent over shuffled mini-batches with the Adadelta update rule, with $l_2$ constraint \cite{hinton2012improving} of 3. We perform early stopping using a subset of the given development dataset. We use beam search of size $10$ to generate the summary.

\paragraph{Baselines}

For the Gigaword dataset, we compare our models with the following abstractive baselines: \textbf{ABS+} \cite{rush2015neural} is a fine tuned version of ABS which uses an attentive CNN encoder and an NNLM decoder, \textbf{Feat2s} \cite{nallapati2016abstractive} is an RNN sequence-to-sequence model with lexical and statistical features in the encoder, \textbf{Luong-NMT} \cite{luong2015effective} is a two-layer LSTM encoder-decoder model, \textbf{RAS-Elman} \cite{chopra2016abstractive} uses an attentive CNN encoder and an Elman RNN decoder, and \textbf{SEASS} \cite{zhou2017selective} uses BiGRU encoders and GRU decoders with selective encoding.
For the CNN dataset, we compare our models with the following extractive and abstractive baselines:
\textbf{Lead-3} is a strong baseline that extracts the first three sentences of the document as summary, \textbf{LexRank} extracts texts using LexRank \cite{erkan2004lexrank}, \textbf{Bi-GRU} is a non-hierarchical one-layer sequence-to-sequence abstractive baseline, \textbf{Distraction-M3} \cite{chen2016distraction} uses a sequence-to-sequence abstractive model with distraction-based networks, and \textbf{GBA} \cite{tan2017abstractive} is a graph-based attentional neural abstractive model.
All baseline results used beam search and are gathered from previous papers. Also, we compare our final model \textbf{\textsc{base}+E2T} with the base model \textbf{\textsc{base}} and some variants of our model (without selective disambiguation, using soft attention).

\section{Results}

\begin{table}[t]
    \centering
    \begin{tabular}{|c|ccc|}
    \hline
    Model & RG-1    & RG-2    & RG-L \\
    \hline
    \hline
    \textsc{base}: s2s+att & 34.14 & 15.44 & 32.47 \\
    \hline
    \textsc{base}+E2T\textsubscript{cnn+sd} & \textbf{37.04} & 16.66 & \textbf{34.93} \\
    \textsc{base}+E2T\textsubscript{rnn+sd} & 36.89 & \textbf{16.86} & 34.74 \\
    \hline
    \textsc{base}+E2T\textsubscript{cnn} & 36.56 & 16.56 & 34.57 \\
    \textsc{base}+E2T\textsubscript{rnn} & 36.52 & 16.21 & 34.32 \\
    \textsc{base}+E2T\textsubscript{cnn+soft} & 36.56 & 16.44 & 34.58 \\
    \textsc{base}+E2T\textsubscript{rnn+soft} & 36.38 & 16.12 & 34.20 \\
    \hline
    \hline
    ABS+ & 29.78 & 11.89 & 26.97 \\
    Feat2s & 32.67 & 15.59 & 30.64 \\
    Luong-NMT & 33.10 & 14.45 & 30.71 \\
    RAS-Elman & 33.78 & 15.97 & 31.15 \\
    SEASS & \textbf{36.15} & \textbf{17.54} & \textbf{33.63} \\
    \hline
    \end{tabular}%
  \caption{Results on the Gigaword dataset using the full-length F1 variants of ROUGE. 
  }
  \label{tab:resgiga}%
\end{table}%

\begin{table}[t]
    \centering
    \begin{tabular}{|c|ccc|}
    \hline
    Model & RG-1    & RG-2    & RG-L \\
    \hline
    \hline
    \textsc{base}: s2s+att & 25.5 & 5.8 & 20.0 \\
    \hline
    \textsc{base}+E2T\textsubscript{cnn+sd} & \textbf{31.9} & \textbf{10.1} & \textbf{23.9} \\
    \textsc{base}+E2T\textsubscript{rnn+sd} & 27.6 & 7.9 & 21.5 \\
    \hline
    \textsc{base}+E2T\textsubscript{cnn} & 26.6 & 7.3 & 20.7 \\
    \textsc{base}+E2T\textsubscript{rnn} & 26.1 & 6.9 & 20.1 \\
    \textsc{base}+E2T\textsubscript{cnn+soft}  & 26.6 & 7.0 & 20.6 \\
    \textsc{base}+E2T\textsubscript{rnn+soft} & 25.0 & 6.7 & 19.8 \\
    \hline
    \hline
    Lead-3 & 26.1 & 9.6 & 17.8 \\
    LexRank & 26.1 & 9.6 & 17.7 \\
    \hline
    Bi-GRU & 19.5 & 5.2 & 15.0 \\
    Distraction-M3 & 27.1 & 8.2 & 18.7 \\
    GBA & \textbf{30.3} & \textbf{9.8} & \textbf{20.0} \\
    \hline
    \end{tabular}%
  \caption{Results on the CNN dataset using the full-length F1 ROUGE metric. 
  }
  \label{tab:rescnn}%
\end{table}%

We report the ROUGE F1 scores for both datasets of all the competing models using ROUGE F1 scores \cite{lin2004rouge}. We report the results on the Gigaword and the CNN dataset in Table \ref{tab:resgiga} and Table \ref{tab:rescnn}, respectively.
In Gigaword dataset where the texts are short, our best model achieves a comparable performance with the current state-of-the-art. 
In CNN dataset where the texts are longer, our best model outperforms all the previous models.
We emphasize that E2T module is easily attachable to better models, and we expect E2T to improve their performance as well.
Overall, E2T achieves a significant improvement over the baseline model \textsc{base}, with at least 2 ROUGE-1 points increase in the Gigaword dataset and 6 ROUGE-1 points increase in the CNN dataset.
In fact, all variants of E2T gain improvements over the baseline, implying that leveraging on linked entities improves the performance of the summarizer. Among the model variants, the CNN-based encoder with selective disambiguation and firm attention performs the best.

Automatic evaluation on the Gigaword dataset shows that the CNN and RNN variants of \textsc{base}+E2T have similar performance.
To break the tie between both models, we also conduct human evaluation on the Gigaword dataset. We instruct two annotators to read the input sentence and rank the competing summaries from first to last according to their relevance and fluency: 
(a) the original summary \textsc{gold}, and
from models (b) \textsc{base}, (c) \textsc{base}+E2T\textsubscript{cnn}, and (d) \textsc{base}+E2T\textsubscript{rnn}. We then compute (i) the proportion of every ranking of each model and (ii) the mean rank of each model. The results are reported in Table \ref{tab:human}.
The model with the best mean rank is \textsc{base}+E2T\textsubscript{cnn}, followed by \textsc{gold}, then by \textsc{base}+E2T\textsubscript{rnn} and \textsc{base}, respectively.
We also perform ANOVA and post-hoc Tukey tests to show that the CNN variant is significantly ($p<0.01$) better than the RNN variant and the base model.
The RNN variant does not perform as well as the CNN variant, contrary to the automatic ROUGE evaluation above.
Interestingly, the CNN variant produces better (but with no significant difference) summaries than the gold summaries. We posit that this is due to the fact that the article title does not correspond to the summary of the first sentence.

\begin{table}
    \small
    \centering
    \begin{tabular}{|c|cccc|c|}
        \hline
        Model & 1st & 2nd & 3rd & 4th & mean \\ \hline
        \textsc{gold} & 0.27 & 0.34 & 0.21 & 0.18 & 2.38 \\ 
        \textsc{base} & 0.14 & 0.15 & 0.28 & \textcolor{red}{0.43} & \textcolor{red}{3.00} \\ 
        \textsc{base}+E2T\textsubscript{rnn} & {0.12} & 0.24 & 0.39 & 0.25 & 2.77 \\ 
        \textsc{base}+E2T\textsubscript{cnn} & \textbf{0.47} & 0.27 & 0.12 & 0.14 & \textbf{1.93} \\ \hline
    \end{tabular}
    \caption{Human evaluations on the Gigaword dataset. Bold-faced values are the best while red-colored values are the worst among the values in the evaluation metric.}
    \label{tab:human}
\end{table}

\begin{table*}[htbp]
  \tiny
  \centering
  \begin{subtable}{\textwidth}
    \begin{tabularx}{\textwidth}{|l|llllllll|}
    \multicolumn{9}{c}{\textbf{Gigaword Dataset Example}} \\
    \hline
    Original & \multicolumn{8}{>{\hsize=\dimexpr8\hsize+14\tabcolsep+6\arrayrulewidth}X|}{western mexico @\textbf{state} @\textbf{jalisco} will host the first edition of the @\textbf{UNK} dollar @\textbf{lorena ochoa} invitation @\textbf{golf} tournament on nov. \#\#-\#\# \#\#\#\# , in @\textbf{guadalajara} @\textbf{country club} , the @\textbf{lorena ochoa} foundation said in a statement on wednesday .} \\ \hline
    Gold  & \multicolumn{8}{>{\hsize=\dimexpr8\hsize+14\tabcolsep+6\arrayrulewidth}X|}{mexico to host lorena ochoa golf tournament in \#\#\#\#} \\ \hline
    Baseline  & \multicolumn{8}{>{\hsize=\dimexpr8\hsize+14\tabcolsep+6\arrayrulewidth}X|}{guadalajara to host ochoa tournament tournament} \\ \hline \hline
    Entities: & \multicolumn{1}{Y}{U.S. state} & \multicolumn{1}{Y}{Jalisco} & \multicolumn{1}{Y}{Unk} & \multicolumn{1}{Y}{Lorena Ochoa} & \multicolumn{1}{Y}{Golf} & \multicolumn{1}{Y}{Guadalajara} & \multicolumn{1}{Y}{Country club} & \multicolumn{1}{c|}{Lorena Ochoa} \\ \hline
    \multirow{2}[0]{*}{Soft} & 
      \multicolumn{1}{Y}{\cellcolor[rgb]{ .984,  .773,  .784}0.083} &
      \multicolumn{1}{Y}{\cellcolor[rgb]{ .984,  .765,  .776}0.086} &
      \multicolumn{1}{Y}{\cellcolor[rgb]{ .98,  .667,  .675}0.124} &
      \multicolumn{1}{Y}{\cellcolor[rgb]{ .984,  .725,  .737}0.101} &
      \multicolumn{1}{Y}{\cellcolor[rgb]{ .984,  .78,  .792}0.080} &
      \multicolumn{1}{Y}{\cellcolor[rgb]{ .98,  .569,  .576}0.161} &
      \multicolumn{1}{Y}{\cellcolor[rgb]{ .976,  .494,  .502}0.189} &
      \multicolumn{1}{c|}{\cellcolor[rgb]{ .976,  .525,  .533}0.177}
      \\
      & \multicolumn{8}{>{\hsize=\dimexpr8\hsize+14\tabcolsep+6\arrayrulewidth}X|}{mexico state \textcolor{red}{guadalajara} to host \textcolor{red}{ochoa} ochoa invitation} \\ \hline
    \multirow{2}[0]{*}{Firm} & 
    \multicolumn{1}{Y}{\cellcolor[rgb]{ .976,  .537,  .545}0.173} &
      \multicolumn{1}{Y}{\cellcolor[rgb]{ .976,  .475,  .482}0.197} &
      \multicolumn{1}{Y}{\cellcolor[rgb]{ .988,  .988,  1}0.000} &
      \multicolumn{1}{Y}{\cellcolor[rgb]{ .976,  .431,  .439}0.213} &
      \multicolumn{1}{Y}{\cellcolor[rgb]{ .976,  .427,  .435}0.215} &
      \multicolumn{1}{Y}{\cellcolor[rgb]{ .988,  .988,  1}0.000} &
      \multicolumn{1}{Y}{\cellcolor[rgb]{ .988,  .988,  1}0.00} &
      \multicolumn{1}{c|}{\cellcolor[rgb]{ .976,  .463,  .471}0.202}
      \\
          & \multicolumn{8}{>{\hsize=\dimexpr8\hsize+14\tabcolsep+6\arrayrulewidth}X|}{mexican state to host first edition of ochoa invitation} \\ \hline
    \end{tabularx}%
    \end{subtable}
    \newline
    \vspace*{2pt}
    \newline
    \centering
    \begin{subtable}{\textwidth}
    \begin{tabularx}{\textwidth}{|l|p{68.3em}|}
    \multicolumn{2}{c}{\textbf{CNN Dataset Example}} \\
    \hline
    Original & URL: \url{http://edition.cnn.com/2015/04/05/politics/netanyahu-iran-deal/index.html} \\ \hline
    Gold  & 
    \makecell[l]{netanyahu says third option is `` standing firm '' to get a better deal . \\
    political sparring continues in u.s. over the deal with iran .}
    \\ \hline
    Baseline  & 
    \makecell[l]{netanyahu says he is a country of `` UNK cheating '' and that it is a country of `` UNK cheating '' \\
    netanyahu says he is a country of `` UNK cheating '' and that `` is a very bad deal '' \\
    he says he says he says the plan is a country of `` UNK cheating '' and that it is a country of `` UNK cheating '' \\
    he says the u.s. is a country of `` UNK cheating '' and that is a country of `` UNK cheating ''}
    \\ \hline 
    \hline
    Soft & 
          \makecell[l]{benjamin netanyahu : `` i think there 's a third alternative , and that is standing firm , '' netanyahu tells cnn . \\
          \textcolor{red}{he says he does not roll back iran 's nuclear ambitions .} \\
          \textcolor{red}{`` it does not roll back iran 's nuclear program . ''}} \\ \hline
    Firm & 
          \makecell[l]{
          new : netanyahu : `` i think there 's a third alternative , and that is standing firm , '' netanyahu says . \\
          \textcolor{red}{obama} 's comments come as democrats and republicans spar over the framework announced last week to lift western sanctions on iran .} \\ \hline
    \end{tabularx}%
    \end{subtable}
  \caption{Examples from Gigaword and CNN datasets and corresponding summaries generated by competing models. The tagged part of text is marked \textbf{bold} and preceded with at sign (@). The red color fill represents the attention scores given to each entity. We only report the attention scores of entities in the Gigaword example for conciseness since there are 80 linked entities in the CNN example.}
  \label{tab:sample}%
\end{table*}%

\begin{table}[ht]
    \tiny
    \centering
    \begin{tabularx}{0.47\textwidth}{|X|c|}
        \hline
        Text & $d$ \\ \hline
        \hline
        \multicolumn{2}{|l|}{Linked entity: \url{https://en.wikipedia.org/wiki/United_States}} \\
        \hline
        \textbf{E1.1}: andy roddick got the better of dmitry tursunov in straight sets on friday , assuring the @\textbf{united states} a \#-\# lead over defending champions russia in the \#\#\#\# davis cup final . & 0.719 \\ \hline
        \textbf{E1.2}: sir alex ferguson revealed friday that david beckham 's move to the @\textbf{united states} had not surprised him because he knew the midfielder would not return to england if he could not come back to manchester united . & 0.086 \\ \hline
        \hline
        \multicolumn{2}{|l|}{Linked entity: \url{https://en.wikipedia.org/wiki/Gold}} \\
        \hline
        \textbf{E2.1}: following is the medal standing at the \#\#th olympic winter games -lrb- tabulated under team , @\textbf{gold} , silver and bronze -rrb- : UNK & 0.862 \\ \hline
        \textbf{E2.2}: @\textbf{gold} opened lower here on monday at \#\#\#.\#\#-\#\#\# .\#\# us dollars an ounce , against friday 's closing rate of \#\#\#.\#\#-\#\#\# .\#\# . & 0.130 \\ \hline
    \end{tabularx}
    \caption{Examples with highest/lowest disambiguation gate $d$ values of two example entities (\textit{United States} and \textit{gold}). The tagged part of text is marked \textbf{bold} and preceded with at sign (@).}
    \label{tab:disambig}
\end{table}

\paragraph{Selective disambiguation of entities}

We show the effectiveness of the selective disambiguation gate $d$ in selecting which entities to disambiguate or not. Table \ref{tab:disambig} shows a total of four different examples of two entities with the highest/lowest $d$ values. In the first example, sentence \textbf{E1.1} contains the entity ``\textit{United States}'' and is linked with the country entity of the same name, however the correct linked entity should be ``\textit{United States Davis Cup team}'', and therefore is given a high $d$ value. On the other hand, sentence \textbf{E1.2} is linked correctly to the country ``\textit{United States}'', and thus is given a low $d$ value.. The second example provides a similar scenario, where sentence \textbf{E2.1} is linked to the entity ``\textit{Gold}'' but should be linked to the entity ``\textit{Gold medal}''. Sentence \textbf{E2.2} is linked correctly to the chemical element. Hence, the former case received a high value $d$ while the latter case received a low $d$ value.

\paragraph{Entities as summary topic}

Finally, we provide one sample for each dataset in Table \ref{tab:sample} for case study, comparing our final model that uses \textbf{firm} attention (\textsc{base}\textsubscript{cnn+sd}), a variant that uses \textbf{soft} attention (\textsc{base}\textsubscript{cnn+soft}), and the \textbf{baseline} model (\textsc{base}). We also show the attention weights of the \textbf{firm} and \textbf{soft} models.

In the Gigaword example, we find three observations.
First, the base model generated a less informative summary, not mentioning ``\textit{mexico state}'' and ``\textit{first edition}''.
Second, the soft model produced a factually wrong summary, saying that ``\textit{guadalajara}'' is a mexican state, while actually it is a city.
Third, the firm model is able to solve the problem by focusing only on the five most important entities, eliminating possible noise such as ``\textit{Unk}'' and less crucial entities such as ``\textit{Country club}''. We can also see the effectiveness of the selective disambiguation in this example, where the entity ``\textit{U.S. state}'' is corrected to mean the entity ``\textit{Mexican state}'' which becomes relevant and is therefore selected.

In the CNN example, we also find that the baseline model generated a very erroneous summary. We argue that this is because the length of the input text is long and the decoder is not guided as to which topics it should focus on.
The soft model generated a much better summary, however it focuses on the wrong topics, specifically on ``\textit{Iran's nuclear program}'', making the summary less general.
A quick read of the original article tells us that the main topic of the article is all about the two political parties arguing over the deal with Iran. However, the entity ``\textit{nuclear}'' appeared a lot in the article, which makes the soft model wrongly focus on the ``\textit{nuclear}'' entity.
The firm model produced the more relevant summary, focusing on the political entities (e.g. ``\textit{republicans}'', ``\textit{democrats}''). This is due to the fact that only the $k=10$ most important elements are attended to create the summary topic vector.

\section{Conclusion}

We proposed to leverage on linked entities to improve the performance of sequence-to-sequence models on neural abstractive summarization task. Linked entities are used to guide the decoding process based on the summary topic and commonsense learned from a knowledge base. We introduced Entity2Topic (E2T), a module that is easily attachable to any 
model using an encoder-decoder framework. E2T applies linked entities into the summarizer by encoding the entities with selective disambiguation and pooling them into one summary topic vector with firm attention mechanism. We showed that by applying E2T to a basic sequence-to-sequence model, we achieve significant improvements over the base model and consequently achieve a comparable performance with more complex summarization models.

\section*{Acknowledgement}
We would like to thank the three anonymous reviewers for their valuable feedback.
This work was supported by Microsoft Research, and Institute for Information communications Technology Promotion (IITP) grant funded by the Korea government (MSIT) (No.2017-0-01778 , Development of Explainable Humanlevel Deep Machine Learning Inference Framework). S. Hwang is a corresponding author.

\bibliography{naaclhlt2018}
\bibliographystyle{acl_natbib}

\end{document}